\pdfoutput=1
\documentclass[11pt, letterpaper, logo, onecolumn, copyright, numbering]{paper_improved}

\usepackage{soul}
\usepackage{xcolor}

\usepackage{booktabs}
\usepackage{enumitem}
\usepackage{array}
\usepackage{colortbl}
\usepackage{xcolor}
\usepackage{caption}
\usepackage{tabularray}
\usepackage[utf8]{inputenc} % allow utf-8 input
\usepackage[T1]{fontenc}    % use 8-bit T1 fonts
\usepackage{hyperref}       % hyperlinks
\usepackage{url}            % simple URL typesetting
\usepackage{booktabs}       % professional-quality tables
\usepackage{amsfonts}       % blackboard math symbols
\usepackage{nicefrac}       % compact symbols for 1/2, etc.
\usepackage{microtype}      % microtypography
\usepackage{xcolor}         % colors
\usepackage{graphicx}

\usepackage{natbib}
\usepackage{verbatim}
\usepackage{inconsolata}
\usepackage{tabularx}
\usepackage{float}
\usepackage{multirow}
\usepackage{listings}
\usepackage{xcolor}
\usepackage{subfigure}
\usepackage{subcaption}
\usepackage{amsmath}
\usepackage{amssymb}
\usepackage{array}
\usepackage{svg}
\usepackage[T1]{fontenc}
\usepackage{wrapfig}

\definecolor{lightgray}{rgb}{0.95, 0.95, 0.95}
\definecolor{darkgray}{rgb}{0.4, 0.4, 0.4}
\definecolor{backcolour}{rgb}{0.95,0.95,0.92}
\definecolor{myblue}{rgb}{0.2, 0.4, 0.8} % Example blue color
\definecolor{mygreen}{rgb}{0.2, 0.6, 0.2} % Example green color

\usepackage{amsmath,amssymb,amsfonts}

\usepackage{CJKutf8}
\usepackage{soul}
\usepackage{url}
\usepackage[utf8]{inputenc}
\usepackage{caption}
\usepackage{graphicx}
\usepackage{xcolor}
\usepackage{amsmath}

\usepackage{amsthm}
\usepackage{booktabs}
\usepackage{algorithm}
\usepackage{latexsym}
\usepackage{graphicx} 
\usepackage{amsmath}
\usepackage{subfigure}
\usepackage{microtype}
\usepackage{algorithmic}
\usepackage[switch]{lineno}
\usepackage[most]{tcolorbox}
\usepackage{alltt}
\definecolor{forestgreen}{rgb}{0.13, 0.55, 0.13}

\tcbset{
    aibox/.style={
        colback=white, 
        colframe=blue!75!black, 
        colbacktitle=blue!85!black, 
        coltitle=white, 
        fonttitle=\bfseries,
        enhanced,  % for shadow and other advanced settings
        drop shadow=black!50!white,  % Shadow
        boxrule=1pt,  % Frame thickness
        boxsep=10pt,  % Inner padding
        left=10pt,  % Padding left
        right=10pt,  % Padding right
        top=6pt,  % Padding top
        bottom=6pt,  % Padding bottom
        title code={\path[tcb fill frame] ([xshift=-10pt]frame.west) -- (frame.north west) -- (frame.north east) -- ([xshift=10pt]frame.east) -- cycle;},  % Title background adjustment
        attach boxed title to top left={xshift=10pt, yshift*=-\tcboxedtitleheight/2},
        boxed title style={boxrule=0pt, frame code={}}  % Title box adjustments
    }
}

\newtcolorbox{AIbox}[2][]{aibox, title=#2, #1}

\let\cite\citep

\title{The Bitter Lesson Learned from 2,000+ Multilingual Benchmarks}
\author{Minghao Wu\textsuperscript{1,2}, Weixuan Wang\textsuperscript{3}, Sinuo Liu\textsuperscript{1,3}, Huifeng Yin\textsuperscript{1,4}, Xintong Wang\textsuperscript{1,5}, Yu Zhao\textsuperscript{1}, Chenyang Lyu\textsuperscript{1}, Longyue Wang\textsuperscript{1}, Weihua Luo\textsuperscript{1}, Kaifu Zhang\textsuperscript{1}\\
\textsuperscript{1} Alibaba International Digital Commerce\\
\textsuperscript{2} Monash University\\
\textsuperscript{3} The University of Edinburgh\\
\textsuperscript{4} Tsinghua University\\
\textsuperscript{5} Universität Hamburg
}

\begin{abstract}

As large language models (LLMs) continue to advance in linguistic capabilities, robust multilingual evaluation has become essential for promoting equitable technological progress. This position paper examines over 2,000 multilingual (non-English) benchmarks from 148 countries, published between 2021 and 2024, to evaluate past, present, and future practices in multilingual benchmarking. Our findings reveal that, despite significant investments amounting to tens of millions of dollars, English remains significantly overrepresented in these benchmarks. Additionally, most benchmarks rely on original language content rather than translations, with the majority sourced from high-resource countries such as China, India, Germany, the UK, and the USA. Furthermore, a comparison of benchmark performance with human judgments highlights notable disparities. STEM-related tasks exhibit strong correlations with human evaluations (0.70 to 0.85), while traditional NLP tasks like question answering (e.g., XQuAD) show much weaker correlations (0.11 to 0.30). Moreover, translating English benchmarks into other languages proves insufficient, as localized benchmarks demonstrate significantly higher alignment with local human judgments (0.68) than their translated counterparts (0.47). This underscores the importance of creating culturally and linguistically tailored benchmarks rather than relying solely on translations. Through this comprehensive analysis, we highlight six key limitations in current multilingual evaluation practices, propose the guiding principles accordingly for effective multilingual benchmarking, and outline five critical research directions to drive progress in the field. Finally, we call for a global collaborative effort to develop human-aligned benchmarks that prioritize real-world applications.

\end{abstract}

\begin{document}

\renewcommand{\tableautorefname}{Table}
\renewcommand{\sectionautorefname}{Section}
\renewcommand{\subsectionautorefname}{Section}
\renewcommand{\subsubsectionautorefname}{Section}
\renewcommand{\figureautorefname}{Figure}
\renewcommand{\equationautorefname}{Equation}
\newcommand{\algorithmautorefname}{Algorithm}
\newcommand{\linenoautorefname}{Line}
\newcommand{\subfigureautorefname}{\figureautorefname}

\maketitle

\begin{figure}[h]
    \centering
    \includegraphics[width=0.93\linewidth]{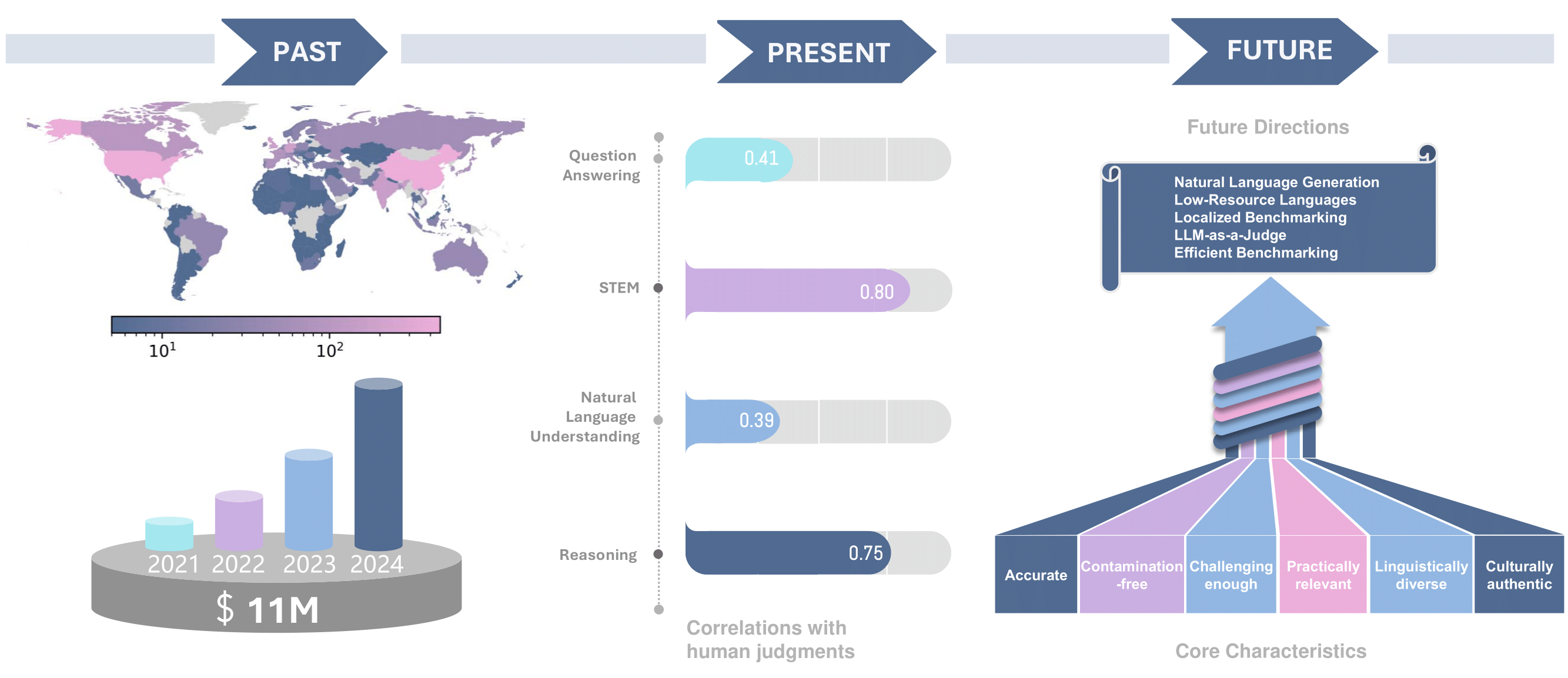}
    \caption{The overview of this work. We examine over 2,000 multilingual (non-English) benchmarks, published between 2021 and 2024, to evaluate past, present, and future practices in multilingual benchmarking. }
    \label{fig:teaser}
\end{figure}

\section{Introduction}

The remarkable capabilities of large language models (LLMs) have transformed natural language processing (NLP), with applications spanning diverse domains and languages worldwide \citep{DBLP:conf/nips/Ouyang0JAWMZASR22, DBLP:conf/iclr/SanhWRBSACSRDBX22,DBLP:journals/corr/abs-2303-08774, DBLP:journals/corr/abs-2302-13971,DBLP:journals/corr/abs-2312-11805,DBLP:journals/corr/abs-2403-08295,DBLP:journals/corr/abs-2412-15115,DBLP:journals/corr/abs-2501-12948}. As these technologies increasingly serve users across linguistic boundaries, robust multilingual evaluation becomes not merely academic but essential \citep{DBLP:journals/corr/abs-2411-11072,DBLP:journals/patterns/QinCZCLLLCY25}. Despite advances in multilingual pre-training, significant disparities persist in how models perform across different languages, writing systems, and cultural contexts. Current evaluation frameworks often fail to capture this diversity, with benchmarks predominantly focused on high-resource languages and evaluation methodologies that inadequately represent real-world use cases in global contexts \citep{hada-etal-2024-large}. This paper presents a comprehensive analysis of multilingual benchmarking practices to understand past approaches, assess present correlations with human judgments, and chart future directions for more equitable, representative, and effective multilingual evaluation of language technologies.

To provide a comprehensive analysis, we first establish a robust dataset of multilingual benchmarks by collecting and annotating papers from the arXiv \texttt{cs.CL} category (2021–2024) (\autoref{sec:scope}). After filtering 370,000 papers through both automated LLM-based screening and expert review, we identify 2,024 relevant studies containing multilingual benchmarks from 148 countries. Our analysis follows a temporal framework organized around three key questions:

\begin{enumerate}
    \item \textbf{PAST: What benchmarks do we currently have? (\autoref{sec:past})}
    We document historical trends, revealing significant disparities in language representation, the evolution of task types, translation methods, and more. Our analysis reveals several key patterns: despite deliberate exclusion of English-only benchmarks, English remains the most represented language, with high-resource languages dominating while low-resource languages are underrepresented. Most benchmarks (61.4\%) use original language content rather than translations, with human translations accounting for just 13.2\%. Discriminative tasks (66.5\%) significantly outnumber generative ones (23.5\%), with text classification remaining dominant while question answering shows dramatic growth following the emergence of LLMs in 2023. Dataset sizes have consistently expanded across all categories over the years, with an estimated cost of \$11 million USD. Most content is sourced from publicly accessible domains, such as news (17\%) and social media (13.3\%), while high-value domains like healthcare and law remain underrepresented. evelopment efforts are concentrated in China, India, Germany, the UK, and the USA, with Europe emphasizing academic research and China and the USA demonstrating stronger academia-industry collaborations.
    
    \item \textbf{PRESENT: What is the current status of multilingual evaluation? (\autoref{sec:present})}
    We present two key insights about multilingual LLM evaluation. First, users across different languages (English, Chinese, French, German, Spanish, and Russian) share remarkably similar interests, with writing tasks dominating (30-45\%), followed by commonsense reasoning and programming tasks, suggesting universal priorities despite linguistic differences. Second, when comparing benchmark performance with human judgments, STEM-related tasks (ARC and MGSM) show consistently stronger correlations across languages (0.70-0.85), while other tasks like XQuAD correlate poorly (0.11-0.30). Importantly, simply translating English benchmarks proves insufficient for robust evaluation, localized benchmarks (like CMMLU for Chinese) show substantially higher correlation with human judgments (0.68) than translated equivalents (0.47 and 0.49), highlighting the critical need for culturally and linguistically authentic evaluation resources.

    \item \textbf{FUTURE: What do we need, and what should we do next? (\autoref{sec:future})}
    Based on our analysis, we outline key principles for effective multilingual benchmarks, emphasizing the need for accurate, contamination-free, challenging, practically relevant, linguistically diverse, and culturally authentic evaluations. It identifies critical research directions, including addressing the imbalance in natural language generation (NLG) tasks, improving representation for low-resource languages, creating localized benchmarks that reflect cultural nuances, leveraging LLMs as multilingual judges while addressing biases, and developing efficient benchmarking methods to manage growing complexity. 
 
\end{enumerate}
Our analysis highlights the existing gaps in multilingual evaluation, including fragmented efforts, limited language coverage, and a mismatch between academic benchmarks and real-world applications. Therefore, we propose a call to action for the NLP community to prioritize the development of multilingual benchmarks that are not only well-aligned with human judgments but also culturally relevant and practically applicable (\autoref{sec:call_to_action}). We emphasize the importance of collaboration among researchers, practitioners, and communities to ensure that language technologies serve all users equitably, regardless of their linguistic background.

Our contributions through this position paper are multifaceted. First, we provide a large-scale and comprehensive analysis to date of multilingual benchmarking trends, documenting historical patterns and identifying critical gaps in language coverage. Second, we quantitatively evaluate how well current benchmarks align with human judgments across multiple languages, offering insights into which evaluation approaches best reflect real-world perceptions of model quality. Third, we propose concrete strategies and a call to action for developing the next generation of multilingual benchmarks, balancing practical constraints with the need for greater linguistic diversity and cultural authenticity. By critically examining existing practices and charting clear directions forward, we aim to catalyze more equitable, representative, and meaningful evaluation methodologies that can better guide the development of truly multilingual language technologies serving the global community.

\section{Related Work}

\paragraph{Multilingual Large Language Models}
Large language models (LLMs) have revolutionized the landscape of natural language processing (NLP) and artificial intelligence (AI) \citep{DBLP:conf/nips/BrownMRSKDNSSAA20,DBLP:conf/nips/Ouyang0JAWMZASR22,DBLP:journals/corr/abs-2212-08073,DBLP:journals/corr/abs-2303-08774,DBLP:journals/corr/abs-2312-11805,DBLP:journals/corr/abs-2408-00118}. Multilingual large language models (MLLMs) also have been developed to address multilingual downstream tasks, extending the capabilities of LLMs to multiple languages \citep{DBLP:journals/corr/abs-2211-05100, DBLP:journals/corr/abs-2405-11804, DBLP:journals/corr/abs-2410-12462}. Early LLMs are typically pre-trained on large-scale English datasets \citep{DBLP:journals/corr/abs-2302-13971, DBLP:journals/corr/abs-2310-06825}. However, recent advancements have made multilingual capabilities a critical feature for modern LLMs. For instance, while Llama-1 models primarily focus on English \citep{DBLP:journals/corr/abs-2302-13971}, the Llama-3.1 series models now support 8 languages \citep{DBLP:journals/corr/abs-2407-21783}. Similarly, Qwen series models initially focus on Chinese and English \citep{DBLP:journals/corr/abs-2309-16609}, but the Qwen 2.5 series models have expanded to support more than 29 languages \citep{DBLP:journals/corr/abs-2412-15115}. To facilitate the development of MLLMs, significant efforts have been made to collect multilingual corpora. A number of multilingual corpora are curated for pre-training \citep{ortiz-suarez-etal-2020-monolingual,DBLP:conf/nips/LaurenconSWAMSW22,nguyen-etal-2024-culturax}. Furthermore, recent works also attempt to collect and synthesize datasets for supervised fine-tuning (SFT) and reinforcement learning from human feedback (RLHF) \citep{DBLP:journals/corr/abs-2305-15011,lai-etal-2023-okapi,singh-etal-2024-aya}. Overall, \citet{DBLP:journals/corr/abs-2411-11072} present a systematic survey on the recent progress of MLLMs.

\paragraph{Multilingual Evaluation}
As large language models (LLMs) continue to advance, there is a growing need for comprehensive and fair evaluation of their capabilities across diverse languages and cultures \citep{DBLP:journals/corr/abs-2310-19736,DBLP:journals/tist/ChangWWWYZCYWWYZCYYX24,wang-etal-2024-retrieval}. Such evaluations are crucial for understanding how well these models perform in multilingual contexts, especially given the linguistic diversity and varying resource availability across languages. Recent research has adopted two primary approaches to multilingual evaluation. The first involves translating existing English evaluation suites into other languages, using either human translators or machine translation systems \citep{DBLP:conf/iclr/ShiSF0SVCTRZ0W23, lai-etal-2023-okapi,DBLP:journals/corr/abs-2412-03304}. The second approach focuses on curating new evaluation suites directly in the target language. For example, inspired by \citet{DBLP:conf/iclr/HendrycksBBZMSS21}, several studies have collected human exam questions from specific regions in the target language to assess LLM performance for local users \citep{koto-etal-2023-large, li-etal-2024-cmmlu, yuksel-etal-2024-turkishmmlu}. In addition to these methods, recent studies have emphasized the inclusion of culturally specific content in their evaluations. For instance, \citet{DBLP:journals/corr/abs-2410-02677} introduced CulturalBench, which assesses LLMs' cultural knowledge across 45 global regions. Similarly, \citet{mohamed-etal-2024-culture} presented ArtELingo-28, a vision-language benchmark spanning 28 languages, with a focus on capturing diverse opinions across languages and cultures. Another notable example is CVQA, introduced by \citet{DBLP:conf/nips/RomeroLWGMPOVBJ24}, which includes culturally driven images and questions from 30 countries across four continents, covering 31 languages and 13 scripts, with a total of 10K questions.

\paragraph{Ours}
In this position paper, we conduct a comprehensive investigation of over 2,000 studies on multilingual evaluation published between 2021 and 2024, following the emergence of large language models (LLMs). We summarize recent trends in the field of multilingual evaluation and analyze whether widely used multilingual benchmarks accurately reflect human judgments. Additionally, we discuss future directions for multilingual evaluation based on our findings. To the best of our knowledge, the work of \citet{yu-etal-2022-beyond}, which surveys 156 multilingual evaluation studies spanning from 2008 to 2021, is the most closely related to our research.

\section{Scope, Collection, and Annotation}
\label{sec:scope}
\begin{table}[t]
\begin{tabular}{lp{0.75\textwidth}}
\toprule
Aspect         & Description                                                                                                          \\ \midrule
\cellcolor{gray!10}Year and Month & \cellcolor{gray!10}The publication year and month of the paper.                                                                         \\
Languages      & The languages covered by the dataset.                                                                                 \\
\cellcolor{gray!10}Task Category  & \cellcolor{gray!10}The task types discussed in the paper (e.g., discriminative, generative, or both).                                   \\
Tasks          & The specific tasks covered (e.g., sentiment analysis, question answering, summarization, etc.).                      \\
\cellcolor{gray!10}Dataset Size   & \cellcolor{gray!10}The approximate size of the dataset, categorized as: $<$100, 100–1K, 1K–10K, 10K–100K, or $>$100K. \\
Affiliation Type & The Affiliation type of the creator of the dataset (e.g. academic, industry, or both).                      \\
\cellcolor{gray!10}Affiliation & \cellcolor{gray!10}The affiliations that create the dataset.  \\
Country    & The countries of the affiliations that create the dataset. \\
\cellcolor{gray!10}Translation    & \cellcolor{gray!10}The method used for dataset translation (e.g., not translated, human translation, Google Translate, etc.).  \\
Domain & The domains of the dataset (e.g., news, social media, etc.). \\
 \bottomrule
\end{tabular}
\caption{
Annotation scheme for the collected paper.
}
\label{tab:annotation}
\end{table}

In this section, we outline our approach to determining the scope of datasets included in our study, detail our collection process from arXiv submissions, and describe our annotation methodology. These steps are critical for ensuring that our benchmark provides a representative and well-documented foundation for evaluating multilingual NLP capabilities.

\paragraph{Scope}
Our work follows the approach of \citet{yu-etal-2022-beyond}, focusing exclusively on labeled datasets in which a system is tasked with generating an output label $y$ from an input text $x$. It is important to note that the output label is not limited to a single categorical value but may also consist of generated text, allowing for the production of more complex outputs. To maintain this focus on clear input-output relationships and ensure that the generated labels remain meaningful and contextually relevant, we deliberately exclude training datasets, unlabeled datasets, machine translation datasets, language identification datasets, and multi-modal datasets from our study. Furthermore, we also exclude the programming languages from our study.

\paragraph{Collection}
In this work, we collect papers under the \texttt{cs.CL} category of arXiv from January 1, 2021, to December 31, 2024 using the arXiv API.\footnote{\url{https://info.arxiv.org/help/api/index.html}} The arXiv API provides programmatic access to metadata and abstracts of papers, enabling efficient data collection. From this process, we initially retrieved a total of 370K papers. To refine the dataset, we utilize \textsc{Qwen2.5-7B-Instruct} to analyze the abstracts of each paper and filter out those irrelevant to our study. Following this automated step, we conduct a manual review to ensure the suitability of each paper for inclusion in our study. This rigorous process resulted in a final dataset of 2,024 papers.

\paragraph{Annotation}
Besides utilizing metadata from the arXiv API, three authors manually annotate the collected papers following the annotation scheme presented in \autoref{tab:annotation}. These authors, each with at least one year of experience in NLP research and proficiency in multiple languages, bring valuable expertise to the annotation process. The additional annotations not only provide new insights into the data but also serve as a valuable resource for future research.

\begin{figure}[t]
\centering
\includegraphics[width=\linewidth]{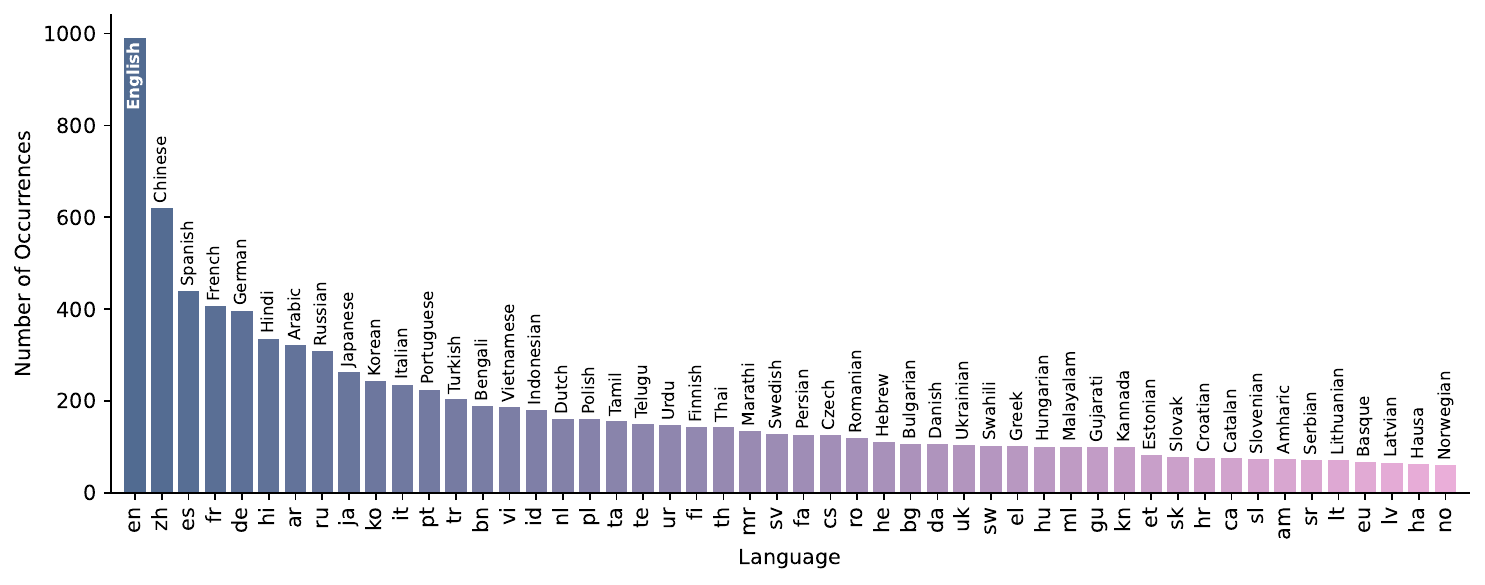}
\caption{
Distribution of the top 50 languages in our multilingual benchmark collection. Although English is deliberately excluded from the collection, it still appears as the most frequent language in the collection. This distribution illustrates the current imbalance in multilingual evaluation benchmarks.
}
\label{fig:top_languages}
\end{figure}

\section{PAST: What Benchmarks Do We Have?}
\label{sec:past}
In this section, we present a comprehensive analysis of the current landscape of multilingual benchmarks based on our paper collection. We examine the distribution of languages, the evolution of task types, translation methods, and more across benchmarks collected from 2021 to 2024. Understanding the existing benchmark ecosystem is crucial for identifying gaps in language coverage, tracking shifts in evaluation focus, and recognizing opportunities for more inclusive and representative benchmark development.

\paragraph{Languages}
\autoref{fig:top_languages} illustrates the distribution of the top 50 languages across our collected benchmarks. Notably, even though we deliberately exclude English benchmarks during the collection process, English still tops the chart, peaking near 1000 occurrences. Similarly, other high-resource languages (HRLs) such as Chinese (zh), Spanish (es), French (fr), and German (de) occupy the leading positions. In contrast, low-resource languages (LRLs) appear much less frequently. This distribution underscores the dominance of high-resource languages within our benchmark collection, while highlighting the challenges in achieving broader linguistic representation.

% \begin{figure}[t]
%     \centering
%     % \includegraphics[width=0.5\linewidth]{figures/translators.pdf}
%     \includegraphics[scale=0.3]{figures/translators1.png}
%     \caption{Translator distribution for the collected papers.}
%     \label{fig:translators}
% \end{figure}

\begin{wrapfigure}{r}{0.5\textwidth}
  \begin{center}
    \includegraphics[width=0.45\textwidth]{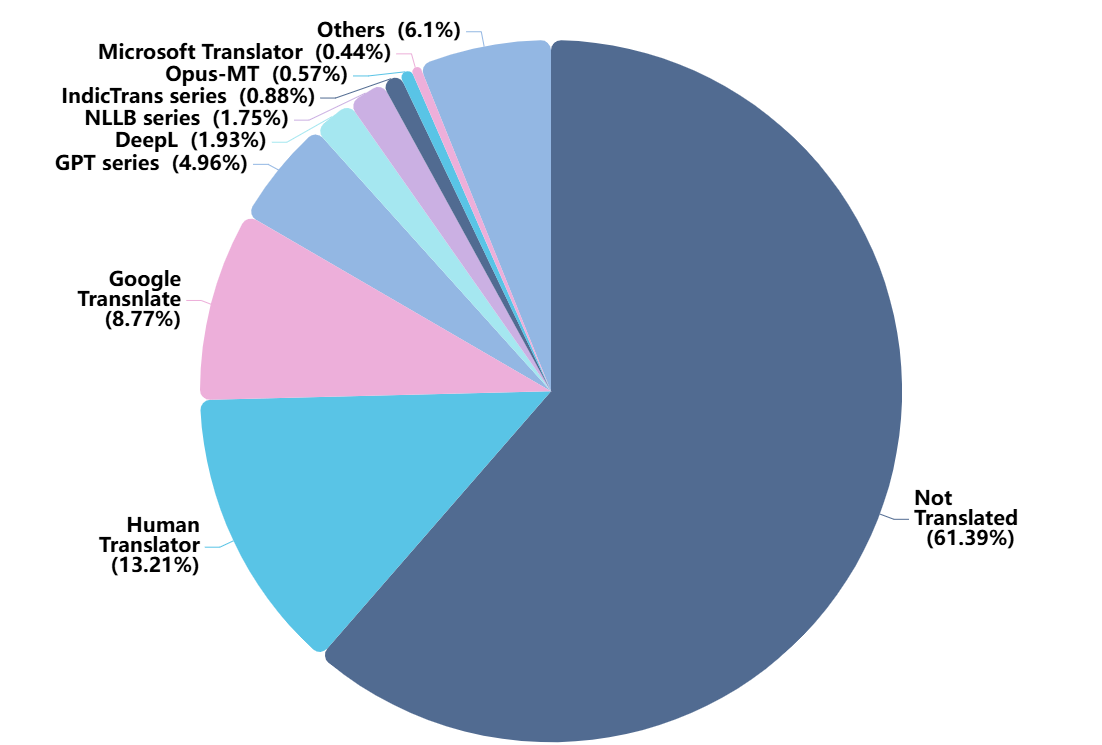}
  \end{center}
  \caption{Distribution of translation methods used in benchmark creation.}
  \label{fig:translations}
\end{wrapfigure}

\paragraph{Translations}
\autoref{fig:translations} illustrates the distribution of translation methods used in benchmark creation. Notably, the majority (61.4\%) of benchmarks are not translated, suggesting they are created in their original languages. Human translations account for 13.2\% of the benchmarks, representing the highest quality but most resource-intensive approach. Among machine translation tools, Google Translate leads with 8.8\%, followed by GPT series models \citep{DBLP:journals/corr/abs-2303-08774} (5.0\%) and DeepL (1.9\%). Specialized translation systems like NLLB series \citep{DBLP:journals/corr/abs-2207-04672} (1.8\%), IndicTrans (0.9\%), Opus-MT \citep{tiedemann-thottingal-2020-opus} (0.6\%), and Microsoft Translator (0.4\%) account for smaller portions. This distribution highlights both the prevalence of native-language benchmark development and the growing role of various machine translation technologies in multilingual benchmark creation.

\begin{figure*}[h!]
\centering          
\subfigure[Tasks]{
\includegraphics[scale=0.6]{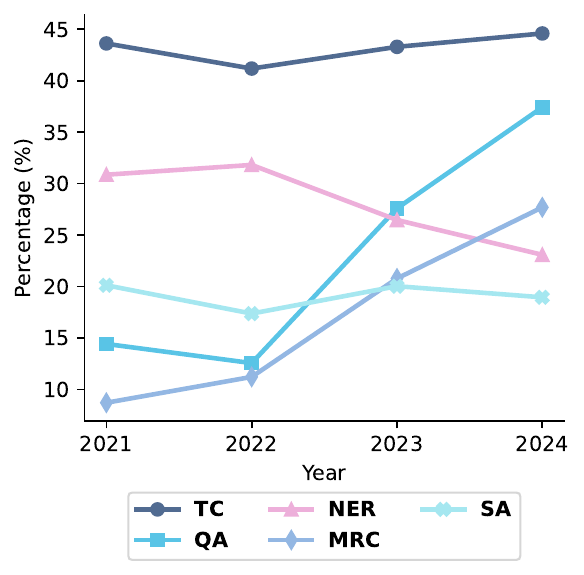}
\label{fig:tasks}
}
\hspace{1cm}
\subfigure[Dataset Sizes]{
\includegraphics[scale=0.6]{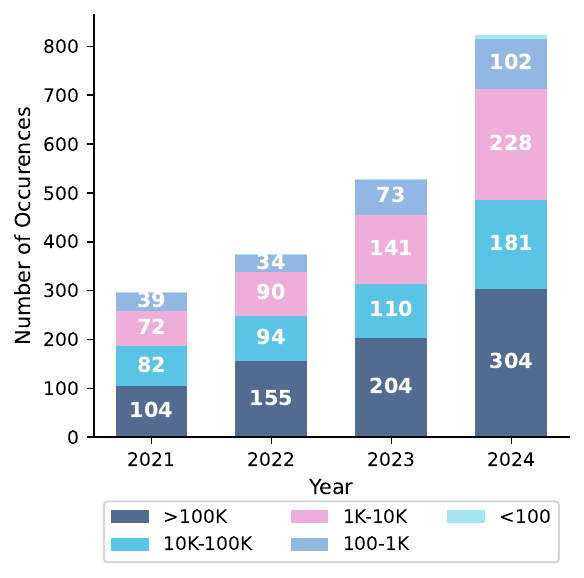}
\label{fig:sizes}
} 
\caption{
(a) Percentage distribution of top 5 tasks in multilingual benchmarks from 2021 to 2024. TC: Text Classification, NER: Named Entity Recognition, SA: Sentiment Analysis, QA: Question Answering, MRC: Machine Reading Comprehension.
(b) Distribution of multilingual benchmarks by dataset size from 2021 to 2024. Categories include very large (>100K examples), large (10K-100K), medium (1K-10K), small (100-1K), and very small (<100) datasets.
}
\label{fig:tasks_and_sizes}
\end{figure*}
\paragraph{Tasks}
In our collected benchmarks, 66.5\% of the papers focus on discriminative tasks, 23.5\% on generative tasks, and 10.0\% on both. \autoref{fig:tasks} shows the percentage distribution of the top 5 tasks from 2021 to 2024. Text classification has consistently been the dominant task. Question answering and Machine reading comprehension have shown dramatic growth. Conversely, named entity recognition has exhibited a declining trend. Sentiment analysis has maintained relatively stable representation. The emergence of large language models (LLMs) in 2023 has significantly reshaped the landscape of multilingual research, with a notable shift towards more complex tasks like question answering.

\paragraph{Dataset Sizes}
\autoref{fig:sizes} illustrates the evolution of dataset sizes in multilingual benchmarks from 2021 to 2024. We observe a consistent upward trend across all size categories, with particularly significant growth in larger datasets. Very large datasets (>100K examples) have nearly tripled from 104 in 2021 to 304 in 2024, while medium (1K-10K) and small datasets (100-1K) have grown from 72 to 228 and 39 to 102 examples, respectively. This trend reflects the growing emphasis on large-scale evaluation resources in the era of foundation models. Assuming each example costs \$0.10 USD on average, these benchmarks have cost more than \$11 million USD from 2021 to 2024.

% \begin{figure}
%     \centering
%     \includegraphics[width=0.5\linewidth]{figures/domains.pdf}
%     \caption{Distribution of domains across multilingual benchmarks in our collection. The chart highlights the prevalence of publicly accessible sources (news, social media, Wikipedia) compared to specialized domains (health, law), reflecting current trends in benchmark creation.}
%     \label{fig:domains}
% \end{figure}

% \begin{figure*}
%     \centering
%     \includegraphics[scale=0.3]{figures/domains1.png}
%     \caption{Distribution of domains across multilingual benchmarks in our collection. The chart highlights the prevalence of publicly accessible sources (news, social media, Wikipedia) compared to specialized domains (health, law), reflecting current trends in benchmark creation.}
%     \label{fig:domains}
% \end{figure*}

\begin{wrapfigure}{r}{0.5\textwidth}
  \begin{center}
    \includegraphics[width=0.45\textwidth]{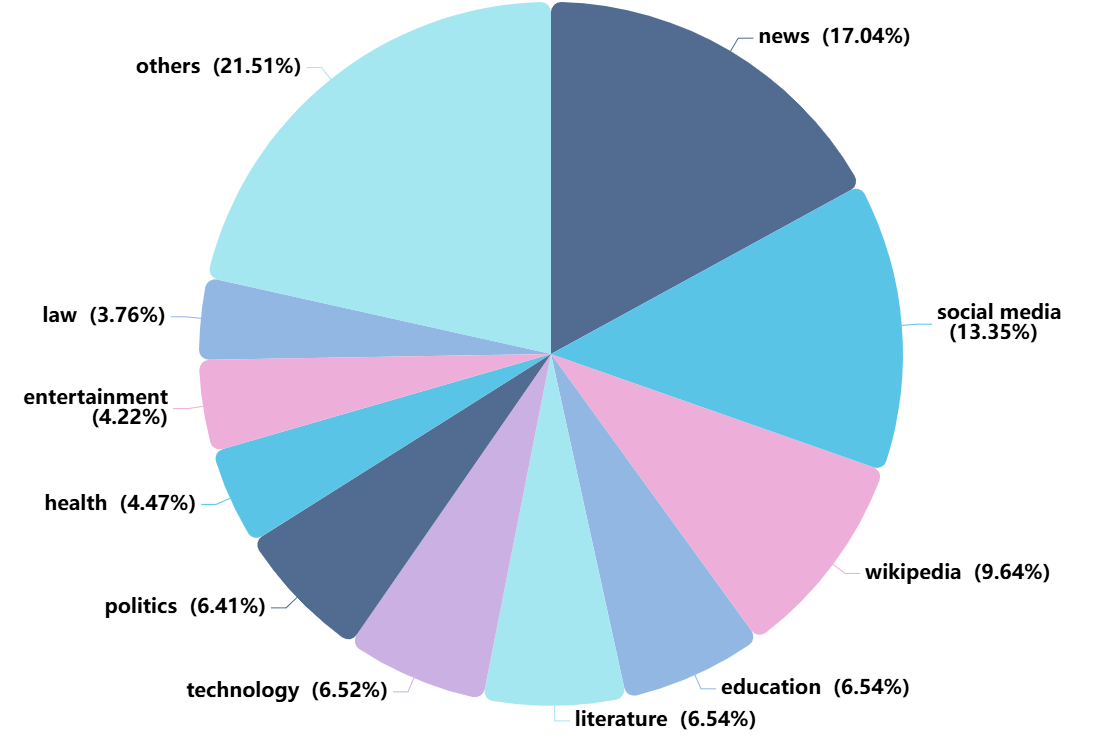}
  \end{center}
  \caption{Distribution of domains across multilingual benchmarks in our collection.}
  \label{fig:domains}
\end{wrapfigure}

\paragraph{Domains}
\autoref{fig:domains} presents the domain distribution across our multilingual benchmark collection. News content constitutes a significant portion at 17.0\% of the benchmarks, followed by social media at 13.3\% and Wikipedia-derived content at 9.6\%. This concentration highlights a clear trend: multilingual benchmarks predominantly leverage publicly accessible sources rather than specialized, high-value domains. Educational and literary domains each represent 6.5\% of the benchmarks, with technology (6.5\%) and politics (6.4\%) following closely behind. More specialized domains like health (4.5\%), entertainment (4.2\%), and law (3.8\%) appear less frequently despite their practical importance. This distribution reflects both the practical convenience of using readily available public data sources and the challenges in obtaining high-quality data from specialized domains, suggesting opportunities for greater representation in these high-value areas.

% year 2021 Counter({'Others': 265, 'USA': 63, 'China': 46, 'India': 25, 'Germany': 22, 'UK': 21})
% year 2022 Counter({'Others': 280, 'USA': 91, 'China': 85, 'UK': 38, 'India': 35, 'Germany': 32})
% year 2023 Counter({'Others': 521, 'China': 145, 'USA': 131, 'Germany': 47, 'India': 38, 'UK': 37})
% year 2024 Counter({'Others': 676, 'China': 209, 'USA': 169, 'Germany': 90, 'UK': 69, 'India': 62})

% \begin{figure}[t]
%     \centering
%     \includegraphics[width=0.5\linewidth]{figures/country.pdf}
%     \caption{Multilingual Benchmarks created by major countries.}
%     \label{fig:country}
% \end{figure}

% \begin{figure*}[t]
% \centering          
% \subfigure[Multilingual Benchmarks created by major countries]{\label{fig:country}\includegraphics[width=0.42 \linewidth]{figures/country1.pdf}} 
% \subfigure[Top 5 instruction categories]{\label{fig:arena}\includegraphics[width=0.48\linewidth]{figures/arena_category_distribution1.pdf}} 
% \caption{(a) . (b).}
% \end{figure*}

\begin{figure*}[t]
\centering          
\subfigure[Country Distributions]{\label{fig:country}\includegraphics[width=0.33\textwidth]{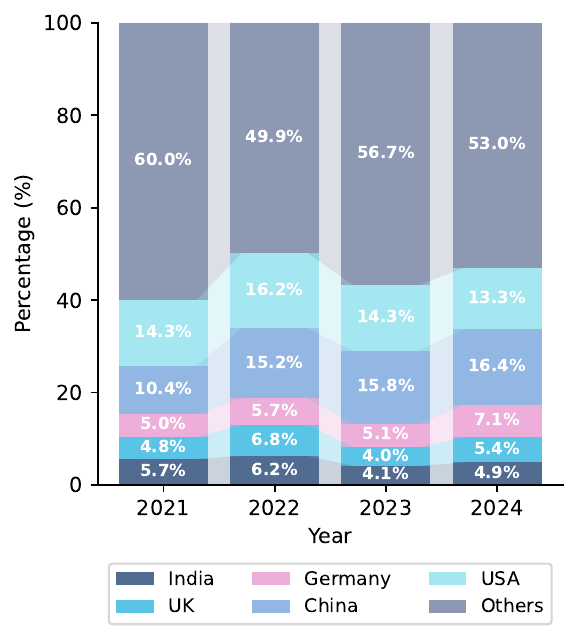}} 
\subfigure[Affiliation Type Distributions of Top 50 countries]{\label{fig:affiliations}\includegraphics[width=0.66\textwidth]{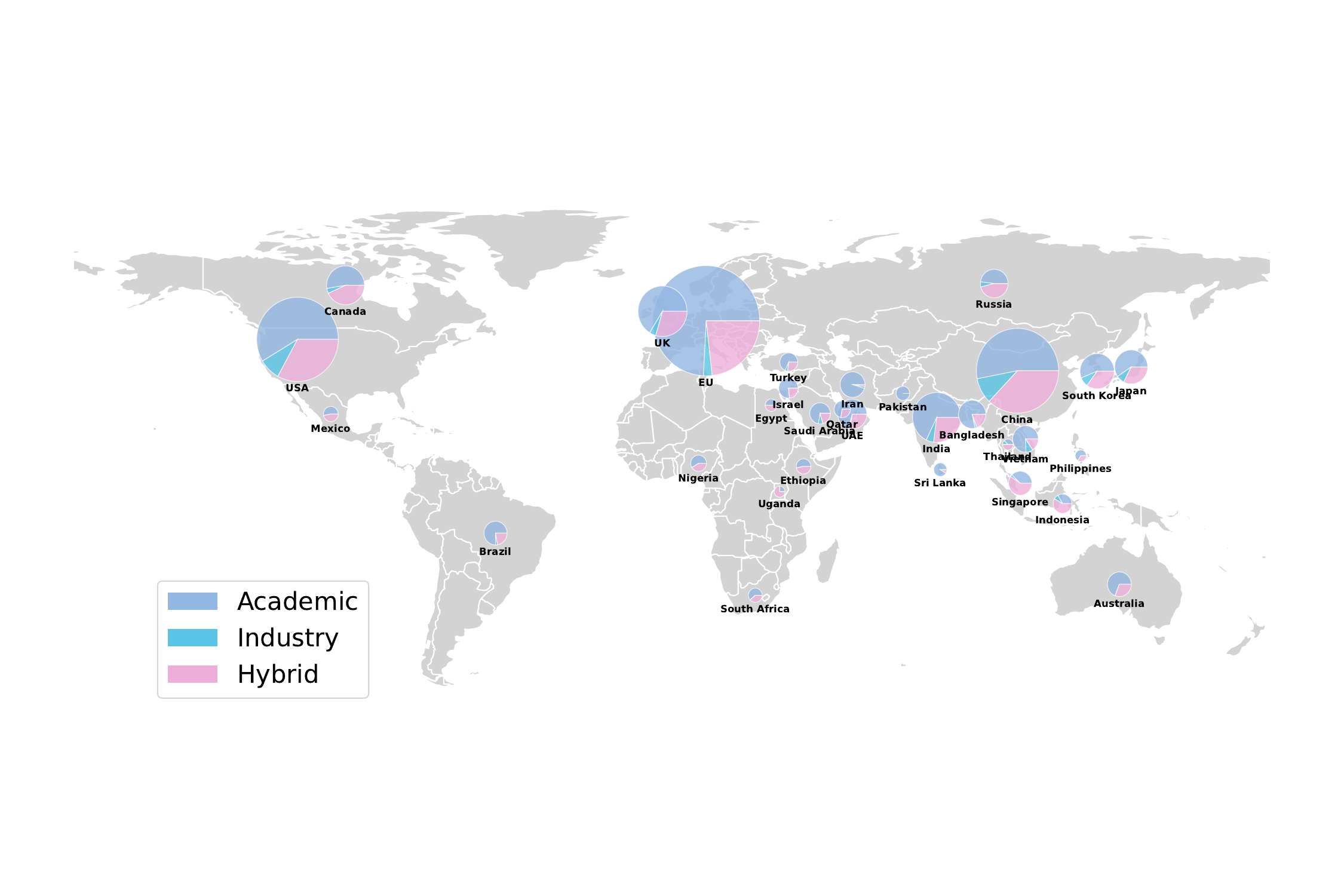}} 
\caption{
(a) Top 5 countries in multilingual benchmark creation from 2021 to 2024. 
(b) Affiliation type distributions of the top 50 countries in multilingual benchmark creation. We merge the countries in European Union (EU) into one category for better visualization.
}
\end{figure*}

\paragraph{Countries and Institutions}
\autoref{fig:country} highlights the countries leading multilingual benchmark development from 2021 to 2024. In this paper, we introduce the term ``G5 countries'' to refer to the China, India, Germany, UK, and USA, which together account for at least 40\% of multilingual benchmark development. Among them, only China shows steady growth from 2021 (10.4\%) to 2024 (16.4\%). \autoref{fig:affiliations} illustrates the institutional distribution across the top 50 countries:\footnote{We use publicly available geographical data for visualization purposes only. The map representation does not imply any political stance or territorial claims. Data Source: \url{https://www.naturalearthdata.com/downloads/110m-cultural-vectors/}} Europe leads with a strong academic focus, while China and the USA feature more balanced academia-industry collaborations. The predominantly academic-driven nature of these benchmarks points to a gap between research and real-world application, suggesting opportunities for greater industry engagement in multilingual benchmark creation.

\section{PRESENT: What is the Current Status of Multilingual Evaluation?}
\label{sec:present}

Understanding the current landscape of multilingual evaluation is essential for developing effective benchmarks and assessment methodologies. In this section, we examine the present state of multilingual evaluation from two critical perspectives: the actual interests and needs of multilingual users (\autoref{sec:interest}), and the alignment between existing multilingual benchmarks and human judgments (\autoref{sec:correlate}). By investigating these dimensions, we aim to identify gaps between user expectations and evaluation practices, and to establish a foundation for more representative and meaningful multilingual benchmarks.

\subsection{What Are the Multilingual Users Interested in?}
\label{sec:interest}

\paragraph{Setup}
To understand the interests of multilingual users, we analyze the distribution of user instructions in Chatbot Arena \citep{DBLP:conf/icml/ChiangZ0ALLZ0JG24} and WildChat \citep{DBLP:conf/iclr/Zhao0HC0D24}. We analyze six languages, including English, Chinese, French, German, Spanish, and Russian, with 10K instructions for each language. We employ \textsc{Qwen2.5-Max} to categorize the instructions. We provide 5 seed categories: \textit{Greeting}, \textit{Writing}, \textit{Translation}, \textit{Math}, and \textit{Programming}, but allow the model to introduce new categories.

\begin{wrapfigure}{r}{0.5\textwidth}
  \begin{center}
    \includegraphics[width=0.45\textwidth]{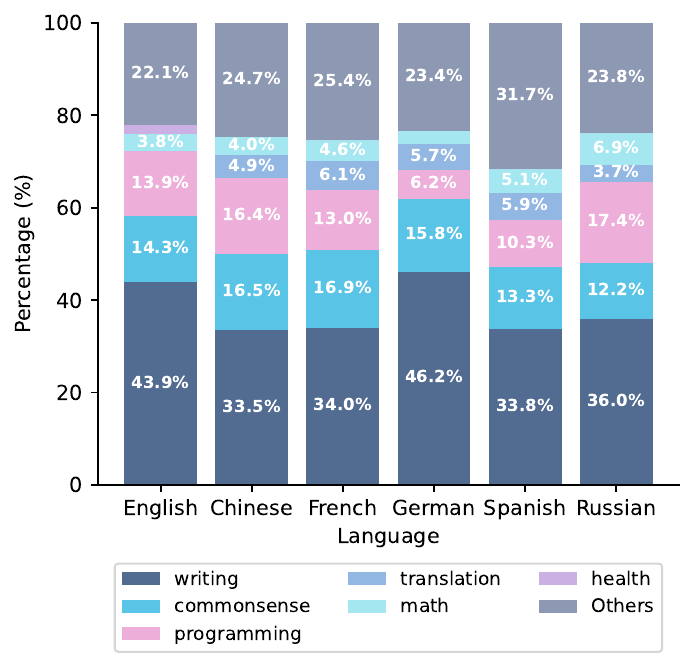}
  \end{center}
  \caption{Distribution of user instruction categories across six languages. We discard the ``Greetings" category, as it is not a task-oriented instruction.}
  \label{fig:arena_category}
\end{wrapfigure}

\paragraph{Users from different countries share common interests.}
We present the distribution of user instructions in \autoref{fig:arena_category}. Our analysis reveals striking similarities in user interests across different languages. Writing tasks dominate user interactions across all six languages, comprising 30-45\% of all instructions. This is followed by commonsense reasoning and programming tasks, which consistently appear among the top three categories in almost all languages. Interestingly, while translation tasks are present in non-English languages (ranging from 5-6\% of instructions), they are understandably absent in English. Mathematical tasks appear consistently across all languages but at lower frequencies (3-7\%). These patterns suggest that despite linguistic and cultural differences, users across different languages primarily use LLMs for similar purposes, with content creation and practical problem-solving being universal priorities. The minor variations observed, such as the higher proportion of programming queries in Russian (17\%) compared to German (6\%), likely reflect differences in professional demographics or educational emphases rather than fundamental divergences in interests. Furthermore, it is important to note that the user instructions are collected from Chatbot Arena and WildChat, which are primarily used for research purposes. Therefore, the distribution of user instructions may not accurately reflect the general population's interests.

\subsection{Do These Benchmarks Correlate Well with Human Judgments?}
\label{sec:correlate}

\paragraph{Setup}
To investigate the correlation between multilingual benchmarks and human judgments, we compare the performance of large language models (LLMs) on multilingual benchmarks with human evaluations collected from Chatbot Arena. Specifically, we evaluate 30 popular LLMs on 8 diverse and widely used multilingual benchmarks: XNLI \citep{conneau-etal-2018-xnli}, ARC \citep{DBLP:journals/corr/abs-1803-05457}, HellaSwag \citep{zellers-etal-2019-hellaswag}, TruthfulQA \citep{lin-etal-2022-truthfulqa}, MMLU \citep{DBLP:conf/iclr/HendrycksBBZMSS21}, GlobalMMLU \citep{DBLP:journals/corr/abs-2412-03304}, XQuAD \citep{artetxe-etal-2020-cross}, and MGSM \citep{DBLP:conf/iclr/ShiSF0SVCTRZ0W23}. For ARC, HellaSwag, TruthfulQA, and MMLU, translations are provided by \citet{lai-etal-2023-okapi} using Google Translate. To assess the alignment between LLM performance and human preferences, we compare the rankings of these models on the benchmarks with their Elo rankings from Chatbot Arena. The Elo rankings, which reflect human judgments, are considered as of the cutoff date of December 30, 2024. The comparison is conducted across five languages: Chinese, French, German, Spanish, and Russian. Spearman's $\rho$ is used as the correlation metric to quantify the relationship between the two ranking systems. We present the list of LLMs used in our evaluation in \autoref{sec:llms}.

\paragraph{STEM-related tasks are more aligned with human judgments.}
As shown in \autoref{tab:correlate}, benchmarks that focus on STEM-related capabilities consistently exhibit stronger correlations with human judgments across all languages. Particularly, ARC and MGSM stand out with remarkably high correlation coefficients. ARC, which tests commonsense and scientific reasoning, shows strong correlations ranging from 0.735 in French to 0.818 in Chinese. Similarly, MGSM, which evaluates mathematical problem-solving abilities, demonstrates exceptionally high correlations across all languages (0.711-0.848). In contrast, natural language inference (XNLI) and question-answering tasks (XQuAD) generally show lower correlations with human preferences. We believe this discrepancy is due to the fact that reasoning capabilities are language agnostic and less affected by translation quality. In contrast, tasks that require understanding of specific linguistic nuances or cultural contexts may be more sensitive to translation errors, leading to weaker correlations with human judgments. Referring to our analysis in \autoref{sec:past}, these findings highlight a concerning trend: while significant financial resources are invested in developing multilingual benchmarks, many of these may fail to align well with human judgments.

\begin{table}[t]
\centering
\small
\setlength{\tabcolsep}{8pt}
\begin{tabular}{llccccc}
\toprule
          & Type & Chinese & French & German & Spanish & Russian \\ \midrule
\multicolumn{7}{l}{\cellcolor{gray!20}\textit{\textbf{Discriminative}}}       \\
XNLI  & Natural Language Inference    & 0.233   & 0.235  & 0.410  & 0.483   & 0.588   \\
ARC &  STEM Question Answering      & 0.818   & 0.735  & 0.767  & \textbf{0.801}   & 0.803   \\
HellaSwag & Commonsense Reasoning & ---     & 0.684  & 0.745  & 0.772   & \textbf{0.811}   \\
TruthfulQA & Question Answering & 0.547   & 0.613  & 0.614  & 0.624   & 0.773   \\
MMLU & Understanding     & 0.473   & 0.398  & 0.371  & 0.345   & 0.303   \\
GlobalMMLU & Understanding & 0.487   & 0.422  & 0.395  & 0.349   & 0.331   \\ \midrule
\multicolumn{7}{l}{\cellcolor{gray!20}\textit{\textbf{Generative}}}           \\
XQuAD & Question Answering     & 0.110   & ---    & 0.301  & 0.225   & 0.154  \\
MGSM &  Mathematics    & \textbf{0.855}   & \textbf{0.814}  & \textbf{0.848}   & 0.798   & 0.711   \\ \bottomrule
\end{tabular}
\caption{
The Spearman's $\rho$ for various benchmarks across 5 languages. The highest correlation for each language is highlighted in \textbf{bold}.
Type indicates the capability type that the benchmark is testing.
}
\label{tab:correlate}
\end{table}

% XNLI: 0.3898
% ARC: 0.7847999999999999
% HellaSwag: 0.753
% TruthfulQA: 0.6342000000000001
% MMLU: 0.378
% GlobalMMLU: 0.3968
% XQuaD: 0.1975
% MGSM: 0.8051999999999999
\paragraph{Translation is NOT all you need.}
The correlation data in \autoref{tab:correlate} clearly demonstrates that simply translating English benchmarks into other languages is insufficient for robust multilingual evaluation. We observe substantial variation in correlation strengths across different languages for the same benchmark. For instance, XNLI shows weak correlation in Chinese and French (0.233 and 0.235) but moderate correlation in Russian (0.588). Similarly, XQuAD exhibits particularly poor correlation in Chinese (0.110) compared to German (0.301). Furthermore, we observe that GlobalMMLU, which contains questions translated by human translators instead of machine translation, shows higher correlations with human judgments compared to the machine-translated MMLU. This finding highlights the importance of high-quality translations.

\paragraph{Localized benchmarks are crucial.}
In addition to the results presented in \autoref{tab:correlate}, we also evaluate these LLMs on CMMLU \citep{li-etal-2024-cmmlu}, which includes authentic exam questions from various Chinese exams. CMMLU demonstrates a correlation of 0.682 with human judgments in Chinese, significantly higher than the correlation of translated MMLU in Chinese (0.473 and 0.487). This finding suggests that translated benchmarks often fail to capture language-specific nuances, cultural contexts, and linguistic features that influence human judgments. It underscores the importance of localized benchmarks specifically designed to capture these nuances and contexts.

\section{FUTURE: What We Need and What We Should Do Next?}
\label{sec:future}

\begin{figure}[t]
    \centering
    \includegraphics[width=0.8\linewidth]{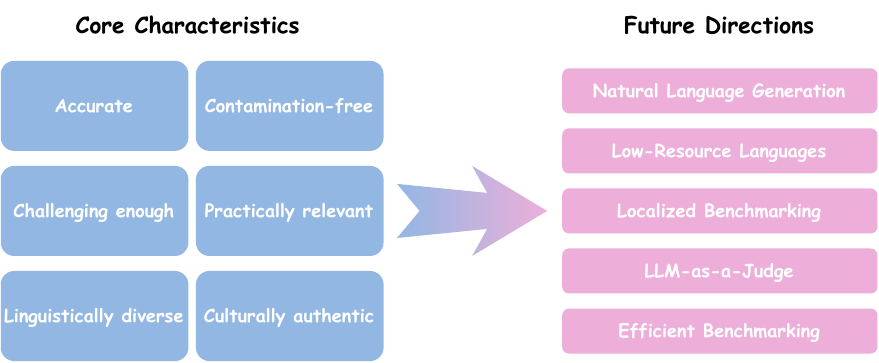}
    \caption{A conceptual framework illustrating the essential characteristics of effective multilingual benchmarks and future research directions for advancing multilingual evaluation.}
    \label{fig:future}
\end{figure}

 This section explores two complementary perspectives, as shown in \autoref{fig:future}. Firstly, we identify essential characteristics of effective multilingual benchmarks (\autoref{sec:character}). Secondly, we propose concrete directions for future research efforts that address persistent gaps in evaluating language models across diverse languages, contexts, and applications (\autoref{sec:directions}). By considering both the ideal benchmark properties and actionable research directions, we aim to chart a path toward more equitable and comprehensive multilingual evaluation.

\subsection{What We Need for Effective Multilingual Benchmarks?}
\label{sec:character}

% \begin{figure}[t]
%     \centering
%     \includegraphics[width=0.8\linewidth]{figures/future.drawio.pdf}
%     \caption{A conceptual framework illustrating the essential characteristics of effective multilingual benchmarks and future research directions for advancing multilingual evaluation.}
%     \label{fig:future}
% \end{figure}

Effective multilingual benchmarks require several key characteristics to meaningfully evaluate LLM capabilities across diverse languages. The development of such benchmarks presents unique challenges beyond those faced in monolingual evaluation settings. Drawing inspiration from \citet{Reiter_2025}, we propose the following key characteristics for good multilingual benchmarks:

\begin{itemize}

    \item \textbf{Accurate:} Benchmark accuracy is foundational to meaningful evaluation. All benchmarks must contain reliable ground truth annotations, properly verified by domain experts. Recent research reveals that even widely adopted benchmarks like MMLU contain numerous errors \citep{gema2025mmlu}, undermining evaluation validity. 
    \item \textbf{Contamination-free:} Benchmark contamination occurs when evaluation data appears in a model's training corpus, leading to inflated performance metrics that misrepresent a model's true capabilities. Recent research even demonstrate that the data contamination in one language can be transferred to another language \citep{yao-etal-2024-data}.
    \item \textbf{Challenging enough:} The performance of recent state-of-the-art models has quickly saturated on widely used benchmarks, with scores approaching or exceeding human performance. As shown in \textsc{Killed-by-LLM},\footnote{\url{https://r0bk.github.io/killedbyllm/}} the average lifespan of a popular benchmark is only 2.6 years before it is not sufficiently challenging. Therefore, multilingual benchmarks must maintain an appropriate difficulty level that can differentiate between models.
    \item \textbf{Practically relevant:} As shown in \autoref{fig:domains} and \autoref{fig:affiliations}, about 70\% of the benchmarks are released by the academic community and these benchmarks are created from the public sources, which may not always reflect real-world applications. Truly useful benchmarks must evaluate capabilities that align with how language technologies are actually used in practice. Without this practical grounding, benchmarks risk optimizing for capabilities that have limited real-world impact, creating a disconnect between research advancements and actual user needs.
    \item \textbf{Linguistically diverse:} Effective multilingual benchmarks must include a diverse set of languages representing different language families, writing systems, and linguistic typologies. Our analysis of language distribution in existing benchmarks (as detailed in \autoref{fig:top_languages}) reveals significant imbalance: English remains the most frequently evaluated language, followed by Chinese, Spanish, and French, despite our intentional exclusion of English-only benchmarks from this study.
    \item \textbf{Culturally authentic:} Multilingual benchmarks must reflect the cultural diversity of the world's languages, ensuring that evaluation tasks are culturally appropriate and authentic. Recent research has highlighted the importance of cultural considerations in benchmark design \citep{DBLP:journals/corr/abs-2402-11548, zhao-etal-2024-worldvaluesbench, DBLP:journals/corr/abs-2410-02677}. Our results also demonstrate that CMMLU \citep{li-etal-2024-cmmlu} aligns better with Chinese users' judgments, compared with the translated MMLU \citep{DBLP:conf/iclr/HendrycksBBZMSS21, DBLP:journals/corr/abs-2412-03304}. 

\end{itemize}

Advancing toward more comprehensive multilingual benchmarks following these principles is essential for ensuring language technologies serve global populations equitably and perform reliably across a wide range of languages.

\subsection{What We Should Do Next?}
\label{sec:directions}

Building on our analysis of necessary characteristics for effective multilingual benchmarks, we now outline five critical research directions.

\paragraph{Natural Language Generation}
While most existing multilingual benchmarks focus on discriminative tasks like classification, question answering, and multiple-choice problems, natural language generation (NLG) capabilities remain significantly underassessed across diverse languages. As discussed in \autoref{sec:past}, about 66\% of the benchmarks are focused on discriminative tasks, while only 23\% of the benchmarks are focused on NLG tasks. This imbalance is particularly concerning as generative applications are increasingly prevalent in real-world deployments of language technologies. 

\paragraph{Low-Resource Languages}
As shown in \autoref{fig:top_languages}, low-resource languages, which lack substantial amounts of digital text data, remain significantly underrepresented in current multilingual benchmarks. This underrepresentation creates a problematic cycle: models perform poorly on these languages, leading researchers to focus on higher-resource languages where improvements are more easily demonstrable, further widening the capability gap. Breaking this cycle requires deliberate effort to develop specialized benchmarks that focus on low-resource languages and their unique linguistic characteristics.

\paragraph{Localized Benchmarking}
Current evaluation approaches often rely on translated content from English or other high-resource languages. The machine-translated evaluation data often preserves source language syntactic structures and cultural references while failing to capture idioms and region-specific knowledge essential to natural language use. As shown in \autoref{sec:correlate}, the localized benchmarks can achieve better alignment with the target language and culture. Recent work has begun addressing these issues by incorporating more diverse cultural perspectives \citep{li-etal-2024-cmmlu, DBLP:journals/corr/abs-2402-11548, zhao-etal-2024-worldvaluesbench, DBLP:journals/corr/abs-2410-02677}, but there remains significant room for benchmarks that assess models on their ability to handle locally meaningful applications.

\paragraph{LLM-as-a-Judge}
Recent research has demonstrated the potential of using LLMs themselves as evaluation tools for assessing the quality of model-generated text in English \citep{DBLP:conf/nips/ZhengC00WZL0LXZ23, DBLP:journals/corr/abs-2404-04475}. This approach offers promising opportunities for multilingual evaluation by extending these techniques across diverse languages and tasks. However, deploying LLMs as judges in multilingual contexts also introduces unique challenges, including potential evaluation biases that mirror the language disparities in the judge models themselves and questions about their cross-lingual evaluation capabilities for low-resource languages. These strengths and challenges highlight the need for further research to develop multilingual LLM judges with more balanced cross-lingual capabilities.

\paragraph{Efficient Benchmarking}
Current benchmarks often include numerous languages and tasks to thoroughly assess model capabilities. The size of benchmarks grows linearly with the number of languages and combinatorially with the number of tasks and evaluation dimensions. As shown in \autoref{fig:sizes}, the size of the benchmarks is growing rapidly over the years. Future research should aim to develop methods for efficient LLM evaluation, such as identifying representative language-task subsets, employing statistical sampling techniques, or using adaptive testing approaches that maintain evaluation quality while reducing computational requirements.

\section{A Call to Action}
\label{sec:call_to_action}

As we conclude our analysis of multilingual benchmarking, we recognize that advancing this field requires not merely technical solutions but a fundamental shift in how we collaborate as a global research community. The challenges identified throughout this paper—from linguistic disparities to cultural authenticity—cannot be addressed by isolated efforts. Therefore, we issue this call to action for researchers, practitioners, and organizations invested in the equitable development of language technologies.

\paragraph{Global Collaboration for Inclusive Benchmarking} 
The creation of truly representative multilingual benchmarks demands collaboration across linguistic, organizational, national, and cultural boundaries. The fragmentation of efforts we observe today leads to significant resource inefficiencies, with multiple teams solving similar problems in isolation. Individual contributions, while valuable, often duplicate work and waste financial and computational resources that could be better allocated through coordinated efforts. We advocate for international research consortia specifically focused on multilingual benchmark development, where expertise across different languages and cultures can be pooled to create more comprehensive evaluation frameworks. Such collaboration would enable more efficient resource allocation while ensuring broader language coverage and deeper linguistic insights than any single organization could achieve alone.

\paragraph{Human-Aligned Evaluation}
Our analysis reveals a critical gap between benchmark performance and meaningful human assessment across different linguistic contexts. Future benchmarking efforts must prioritize alignment with diverse human judgments, recognizing that language proficiency is ultimately about human communication and understanding. This requires engaging native speakers and linguistic experts from varied backgrounds in benchmark design, validation, and interpretation. Beyond mere translation of existing benchmarks, we need evaluation paradigms that capture language-specific nuances, cultural references, and pragmatic understanding that reflect how humans actually judge language quality in different cultural contexts. The disconnect between benchmark scores and human perception undermines the practical utility of our evaluation methods and must be addressed through systematic incorporation of human feedback.

\paragraph{Application-Oriented Benchmarking}
The academic nature of many existing benchmarks creates a concerning divide between what we measure and what matters in real-world deployments. Future multilingual benchmarks should be designed to reflect practical applications across diverse contexts, such as education, healthcare, commerce, and more. This requires engaging stakeholders beyond academic research, including industry practitioners, public service providers, and community organizations across different regions. By grounding benchmark tasks in concrete use cases that vary by linguistic and cultural context, we can ensure our evaluation methods drive improvements in language technologies that truly serve diverse global communities. Such application-oriented benchmarks would help bridge the gap between research advancements and meaningful impact in multilingual settings.

The path forward for multilingual benchmarking requires commitment from all stakeholders in the language technology ecosystem. By fostering collaboration across boundaries, centering human judgment in our evaluation methods, and focusing on real-world applications, we can develop benchmarking approaches that drive equitable progress in language technologies. The global nature of communication demands nothing less than a global effort to ensure language models serve all communities fairly and effectively. We invite researchers, practitioners, and organizations to join this collaborative effort toward more inclusive, representative, and meaningful multilingual evaluation.

\section{Conclusion}

In this position paper, we present a comprehensive analysis of multilingual benchmarking practices by systematically examining over 2,000 studies. Our findings uncover persistent disparities in language representation, evolving task types, dataset sizes, and other critical factors. Through a present-focused investigation, we identify user interests across different languages and highlight significant gaps between benchmark scores and actual human preferences, particularly in translation-based evaluations. Our analysis underscores six key limitations in current multilingual evaluation practices and proposes guiding principles for effective multilingual benchmarking. Additionally, we outline five critical research directions to advance the field. Finally, we advocate for a global collaborative effort to develop human-aligned benchmarks that prioritize real-world applications and address the diverse needs of multilingual users.

\bibliography{references}

% \appendix % This command starts the appendix
% \input{6-Appendix}

\clearpage
\appendix
\section{LLMs in Evaluations}
\label{sec:llms}
\begin{table}[t]
\centering
\small
\begin{tabular}{lccccc}
\toprule
Model                                & Chinese & French           & German           & Spanish          & Russian \\ \midrule
google/gemma-1.1-7b-it               & 1118.6  & 1018.5           & 1049.9           & 1052.1           & 1076.6  \\
CohereForAI/aya-expanse-32b          & 1267.1  & 1199.8           & 1196.6           & 1199.7           & 1249.8  \\
google/gemma-7b-it                   & 1095.5  & \phantom{0}979.7 & \phantom{0}978.9 & \phantom{0}983.3 & 1014.7  \\
meta-llama/Llama-3.1-8B-Instruct     & 1211.2  & 1142.8           & 1138.8           & 1176.5           & 1187.2  \\
meta-llama/Llama-2-7b-chat-hf        & 1031.8  & \phantom{0}925.1 & \phantom{0}956.4 & \phantom{0}989.9 & 1015.4  \\
microsoft/Phi-3-small-8k-instruct    & 1122.9  & 1091.9           & 1075.4           & 1110.6           & 1138.4  \\
Qwen/Qwen2.5-Coder-32B-Instruct      & 1277.2  & 1182.5           & 1192.8           & 1219.7           & 1250.3  \\
meta-llama/Meta-Llama-3-8B-Instruct  & 1135.2  & 1113.1           & 1101.6           & 1174.5           & 1138.8  \\
ibm-granite/granite-3.0-8b-instruct  & 1130.5  & 1027.3           & \phantom{0}983.2 & 1034.0           & 1102.5  \\
microsoft/Phi-3-medium-4k-instruct   & 1165.1  & 1070.4           & 1100.7           & 1098.7           & 1169.8  \\
google/gemma-2-27b-it                & 1278.8  & 1190.1           & 1206.0           & 1223.9           & 1255.9  \\
google/gemma-1.1-2b-it               & 1076.4  & \phantom{0}963.9 & \phantom{0}947.8            & \phantom{0}991.4            & 1020.1  \\
microsoft/Phi-3-mini-128k-instruct   & 1076.3  & \phantom{0}994.2            & 1009.0           & 1056.1           & 1039.0  \\
meta-llama/Llama-3.2-1B-Instruct     & 1023.0  & 1021.4           & 1010.9           & 1030.2           & \phantom{0}972.9   \\
CohereForAI/aya-expanse-8b           & 1241.2  & 1166.1           & 1180.3           & 1161.4           & 1228.4  \\
meta-llama/Llama-2-13b-chat-hf       & 1055.3  & \phantom{0}992.3            & \phantom{0}998.2            & 1076.9           & 1075.9  \\
meta-llama/Llama-3.2-3B-Instruct     & 1084.7  & 1031.0           & 1053.6           & 1095.7           & \phantom{0}984.0   \\
google/gemma-2-2b-it                 & 1190.3  & 1129.3           & 1105.1           & 1144.6           & 1142.1  \\
HuggingFaceH4/zephyr-7b-beta         & 1017.9  & \phantom{0}989.5            & \phantom{0}975.5            & 1040.4           & 1067.1  \\
microsoft/Phi-3-mini-4k-instruct     & 1081.8  & 1033.2           & 1038.5           & 1094.3           & 1056.5  \\
google/gemma-2b-it                   & 1049.3  & \phantom{0}852.2            & \phantom{0}909.9            & \phantom{0}985.9            & \phantom{0}964.9   \\
mistralai/Mistral-7B-Instruct-v0.2   & 1068.4  & \phantom{0}983.9            & \phantom{0}979.2            & 1025.7           & 1045.8  \\
HuggingFaceTB/SmolLM2-1.7B-Instruct  & 1106.5  & 1001.9           & \phantom{0}948.5            & \phantom{0}941.4            & 1033.0  \\
Qwen/Qwen1.5-14B-Chat                & 1202.6  & 1068.4           & 1042.3           & 1079.2           & 1073.3  \\
google/gemma-2-9b-it                 & 1243.3  & 1142.9           & 1180.2           & 1199.7           & 1220.1  \\
Qwen/Qwen1.5-4B-Chat                 & 1083.6  & \phantom{0}929.8            & \phantom{0}904.9            & 1013.6           & \phantom{0}977.5   \\
mistralai/Ministral-8B-Instruct-2410 & 1256.8  & 1133.4           & 1128.0           & 1131.1           & 1222.9  \\
ibm-granite/granite-3.0-2b-instruct  & 1130.3  & 1003.1           & \phantom{0}988.4            & 1037.0           & 1081.4  \\
Qwen/Qwen1.5-7B-Chat                 & 1196.1  & 1017.3           & 1022.4           & 1012.4           & 1035.8  \\
allenai/OLMo-7B-Instruct             & 1071.7  & \phantom{0}879.6            & \phantom{0}885.3            & \phantom{0}975.9            & \phantom{0}970.9  \\ \bottomrule
\end{tabular}
\caption{LLMs used for evaluation and their Elo scores on 5 languages up to December 30, 2024.}
\label{tab:llms}
\end{table}
In this work, we evaluate 30 LLMs on 8 multilingual benchmarks across 5 languages. We present all the LLMs used in this work in \autoref{tab:llms}. 

\end{document}